\documentclass[10pt,twocolumn,letterpaper]{article}

\usepackage[pagenumbers]{cvpr}              
\usepackage[table,dvipsnames]{xcolor}
\usepackage{multirow}
\usepackage{subcaption}

\usepackage[dvipsnames]{xcolor}
\usepackage{comment}

\definecolor{cvprblue}{rgb}{0.21,0.49,0.74}
\usepackage[pagebackref,breaklinks,colorlinks,citecolor=cvprblue]{hyperref}
\usepackage{float}

\newcommand{\specialfootnotetext}[1]{
  \begingroup
  \def\thefootnote{\fnsymbol{footnote}} 
  \footnotetext[0]{#1} 
  \endgroup
}

\def\paperID{5760} 
\def\confName{CVPR}
\def\confYear{2024}

\title{FreeControl: Training-Free Spatial Control of Any Text-to-Image Diffusion Model with Any Condition\\[0.02in]
{\large \href{https://genforce.github.io/freecontrol/}{https://genforce.github.io/freecontrol/}}
\vspace{-1em}}
\author{Sicheng Mo\textsuperscript{\dag *}, Fangzhou Mu\textsuperscript{\S *}, Kuan Heng Lin\textsuperscript{\dag}, Yanli Liu\textsuperscript{\ddag}, Bochen Guan\textsuperscript{\ddag}, Yin Li\textsuperscript{\S}, Bolei Zhou\textsuperscript{\dag} \\ 
{\small \textsuperscript{\dag}University of California, Los Angeles, \textsuperscript{\S}University of Wisconsin-Madison, \textsuperscript{\ddag}Innopeak Technology, Inc}
}

\begin{document}

\twocolumn[{
\renewcommand\twocolumn[1][]{#1}
\maketitle

\vspace{-3em}
\begin{center}
    \centering
    \captionsetup{type=figure}
    \includegraphics[width=0.95\textwidth]{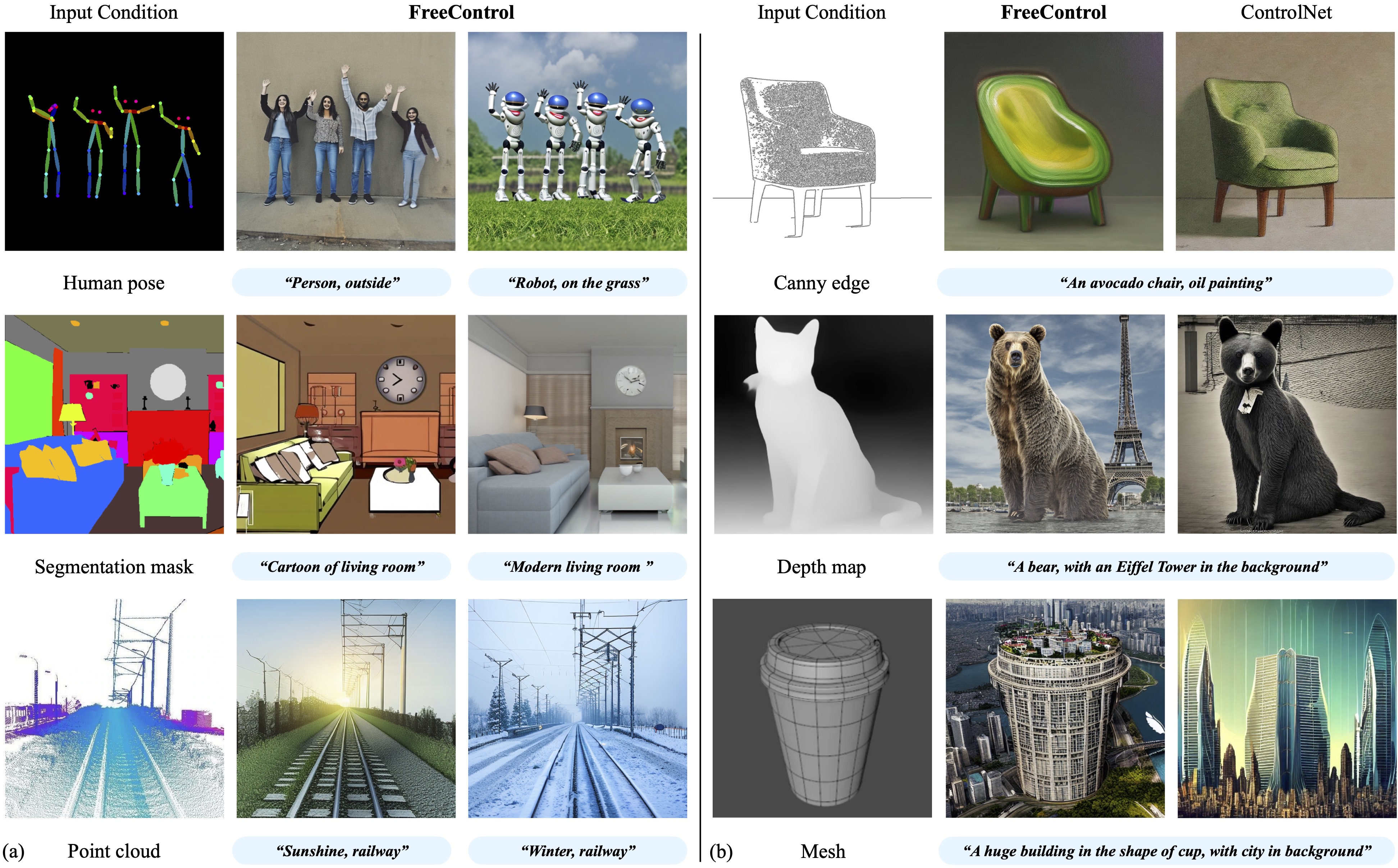}
    \captionof{figure}{{\bf Training-free conditional control of Stable Diffusion.} (a) FreeControl enables zero-shot control of pretrained text-to-image diffusion models given an input condition image in any modality. (b) Compared to ControlNet~\cite{zhang2023controlNet}, FreeControl achieves a good balance between spatial and image-text alignment when facing a conflict between the guidance image and text description. Further, it supports condition types (\eg, 2D projections of point clouds and meshes in the borrow row) for which constructing training pairs is difficult.
    }
    \label{fig:teaser}
\end{center}
}]
\specialfootnotetext{* indicates equal contribution}
\begin{abstract}
Recent approaches such as ControlNet~\cite{zhang2023controlNet} offer users fine-grained spatial control over text-to-image (T2I) diffusion models. However, auxiliary modules have to be trained for each type of spatial condition, model architecture, and checkpoint, putting them at odds with the diverse intents and preferences a human designer would like to convey to the AI models during the content creation process. 
In this work, we present FreeControl, a training-free approach for controllable T2I generation that supports multiple conditions, architectures, and checkpoints simultaneously. 
FreeControl designs structure guidance to facilitate the structure alignment with a guidance image, and appearance guidance to enable the appearance sharing between images generated using the same seed.
Extensive qualitative and quantitative experiments demonstrate the superior performance of FreeControl across a variety of pre-trained T2I models. In particular, FreeControl facilitates convenient training-free control over many different architectures and checkpoints, allows the challenging input conditions on which most of the existing training-free methods fail, and achieves competitive synthesis quality with training-based approaches. 
\end{abstract}\vspace{-0.5em}

\section{Introduction}
\label{sec:intro}

Text-to-image (T2I) diffusion models~\cite{ramesh2022dalle2,midjourney} have achieved tremendous success in high-quality image synthesis, yet a text description alone is far from enough for humans to convey their preferences and intents for content creation. Recent advances such as ControlNet~\cite{zhang2023controlNet} enable spatial control of pretrained T2I diffusion models, allowing users to specify the desired image composition by providing a guidance image (\eg, depth map, human pose) alongside the text description. Despite their superior generation results, these methods~\cite{zhang2023controlNet,li2023gligen,mou2023t2i,zhao2023uni,voynov2023sketch,avrahami2023spatext} require training an additional module specific to each type of spatial condition. Considering the large space of control signals, constantly evolving model architectures, and a growing number of customized model checkpoints (\eg, Stable Diffusion~\cite{rombach2022ldm} fine-tuned for Disney characters or user-specified objects~\cite{ruiz2023dreambooth,hu2022lora}), this repetitive training on every new model and condition type is costly and wasteful, if not infeasible.

Beyond high training cost and poor scalability, controllable T2I diffusion methods face drawbacks that stem from their training scheme: they are trained to output a target image given a spatially-aligned control condition derived automatically from the same image using an off-the-shelf vision model (\eg, MiDaS~\cite{ranftl2020midas} for depth maps, OpenPose~\cite{cao2017openpose} for human poses). This limits the use of many desired control signals that are difficult to infer from images (\eg, mesh, point cloud). Further, the trained models tend to prioritize spatial conditions over text descriptions, likely because the close spatial alignment of input-output image pairs exposes a shortcut. This is illustrated in Figure~\ref{fig:teaser}(b), where there is a conflict between the guidance image and text prompt (\eg, an edge map of a sofa chair \vs ``an avocado chair'').

To address these limitations, we present FreeControl, a versatile training-free method for controllable T2I diffusion. Our key motivation is that feature maps in T2I models during the generation process already capture the spatial structure and local appearance described in the input text. By modeling the subspace of these features, we can effectively steer the generation process towards a similar structure expressed in the guidance image, while preserving the appearance of the concept in the input text. To this end, FreeControl combines an analysis stage and a synthesis stage. In the analysis stage, FreeControl queries a T2I model to generate as few as one seed image and then constructs a linear feature subspace from the generated images. In the synthesis stage, FreeControl employs guidance in the subspace to facilitate structure alignment with a guidance image, as well as appearance alignment between images generated with and without control.

FreeControl offers a significant advantage over training-based methods by eliminating the need for additional training on a pretrained T2I model, while adeptly adhering to concepts outlined in the text description. It supports a wide array of control conditions, model architectures and customized checkpoints, achieves high-quality image generation with robust controllability in comparison to prior training-free methods~\cite{meng2021sdedit,hertz2022prompt-to-prompt,tumanyan2023plug-and-play,parmar2023pix2pix-zero}, and can be readily adapted for text-guided image-to-image translation. We conduct extensive qualitative and quantitative experiments and demonstrate the superior performance of our method. Notably, FreeControl excels at challenging control conditions on which prior training-free methods fail. In the meantime, it attains competitive image synthesis quality with training-based methods while providing stronger image-text alignment and supporting a broader set of control signals.

\smallskip
\noindent \textbf{Contributions}. 
(1) We present FreeControl, a novel method for training-free controllable T2I generation via modeling the linear subspace of intermediate diffusion features and employing guidance in this subspace during the generation process.
(2) Our method presents the first universal training-free solution that supports multiple control conditions (sketch, normal map, depth map, edge map, human pose, segmentation mask, natural image and beyond), model architectures (\eg, SD 1.5, 2.1, and SD-XL 1.0), and customized checkpoints (\eg, using DreamBooth~\cite{ruiz2023dreambooth} and LoRA~\cite{hu2022lora}). (3) Our method demonstrates superior results in comparison to previous training-free methods (\eg, Plug-and-Play~\cite{tumanyan2023plug-and-play}) and achieves competitive performance with prior training-based approaches (\eg, ControlNet~\cite{zhang2023controlNet}).

\section{Related Work}
\label{sec:related_work}

\noindent\textbf{Text-to-image diffusion.}
Diffusion models~\cite{sohl2015thermo,ho2020ddpm,song2020score} bring a recent breakthrough in text-to-image (T2I) generation. 
T2I diffusion models formulate image generation as an iterative denoising task guided by a text prompt. Denoising is conditioned on textual embeddings produced by language encoders~\cite{raffel2020t5,radford2021clip} and is performed either in pixel space~\cite{nichol2021glide,ramesh2022dalle2,saharia2022imagen,balaji2022ediffi} or latent space~\cite{rombach2022ldm,gu2022vq-diffusion,podell2023sdxl}, followed by cascaded super-resolution~\cite{ho2022cascaded} or latent-to-image decoding~\cite{esser2021taming} for high-resolution image synthesis. Several recent works show that the internal representations of T2I diffusion models capture mid/high-level semantic concepts, and thus can be repurposed for image recognition tasks~\cite{xu2023open,li2023diffusion}. Our work builds on this intuition and explores the feature space of T2I models to guide the generation process. 


\smallskip
\noindent\textbf{Controllable T2I diffusion.}
It is challenging to convey human preferences and intents through text description alone. Several methods thus instrument pre-trained T2I models to take an additional input condition by learning auxiliary modules on paired data~\cite{zhang2023controlNet,li2023gligen,mou2023t2i,zhao2023uni,voynov2023sketch,avrahami2023spatext}. One significant drawback of this training-based approach is the cost of repeated training for every control signal type, model architecture, and model checkpoint. On the other hand, training-free methods leverage attention weights and features inside a pre-trained T2I model for the control of object size, shape, appearance and location~\cite{patashnik2023localizing,cao2023masactrl,xie2023boxdiff,epstein2023selfguidance,ge2023expressive}. However, these methods only take coarse conditions such as bounding boxes to achieve precise control over object pose and scene composition. Different from all the prior works, FreeControl is a training-free approach to controllable T2I diffusion that supports any spatial conditions, model architectures, and checkpoints within a unified framework.

\smallskip
\noindent\textbf{Image-to-image translation with T2I diffusion.}
Controlling T2I diffusion becomes an image-to-image translation (I2I) task~\cite{isola2017im2im} when the control signal is an image. I2I methods map an image from its source domain to a target domain while preserving the underlying structure~\cite{isola2017im2im,park2019spade,saharia2022palette}. T2I diffusion enables I2I methods to specify target domains using text. Text-driven I2I is often posed as conditional generation~\cite{zhang2023controlNet,mou2023t2i,zhao2023uni,brooks2023instructpix2pix,kawar2023imagic,zhang2023sine}. These methods finetune a pretrained model to condition it on an input image. Alternatively, recent training-free methods perform zero-shot image translation~\cite{meng2021sdedit,hertz2022prompt-to-prompt,tumanyan2023plug-and-play,parmar2023pix2pix-zero} and is most relevant to our work. This is achieved by inverting the input image~\cite{song2020ddim,mokady2023null,wallace2023edict}, followed by manipulating the attention weights and features throughout the diffusion process. A key limitation of these methods is they require the input to have rich textures, and hence they fall short when converting abstract layouts (\eg depth) to realistic image. By contrast, our method attends to \emph{semantic} image structure by decomposing features into principal components, thereby it supports a wide range of modalities as layout specifications.

\smallskip
\noindent\textbf{Customized T2I diffusion.}
Model customization is a key use case of T2I diffusion in visual content creation. By fine-tuning a pretrained model on images of custom objects or styles, several methods~\cite{ruiz2023dreambooth,gal2022ti,kumari2023customdiffusion,avrahami2023break-a-scene} bind a dedicated token to each concept and insert them in text prompts for customized generation. Amid the growing number of customized models being built and shared by content creators~\cite{huggingface,civitai}, FreeControl offers a \emph{scalable} framework for zero-shot control of any model with any spatial condition. 
\section{Preliminary}
\label{sec:prelimilary}

\begin{figure}
\centering
\includegraphics[width=\linewidth]{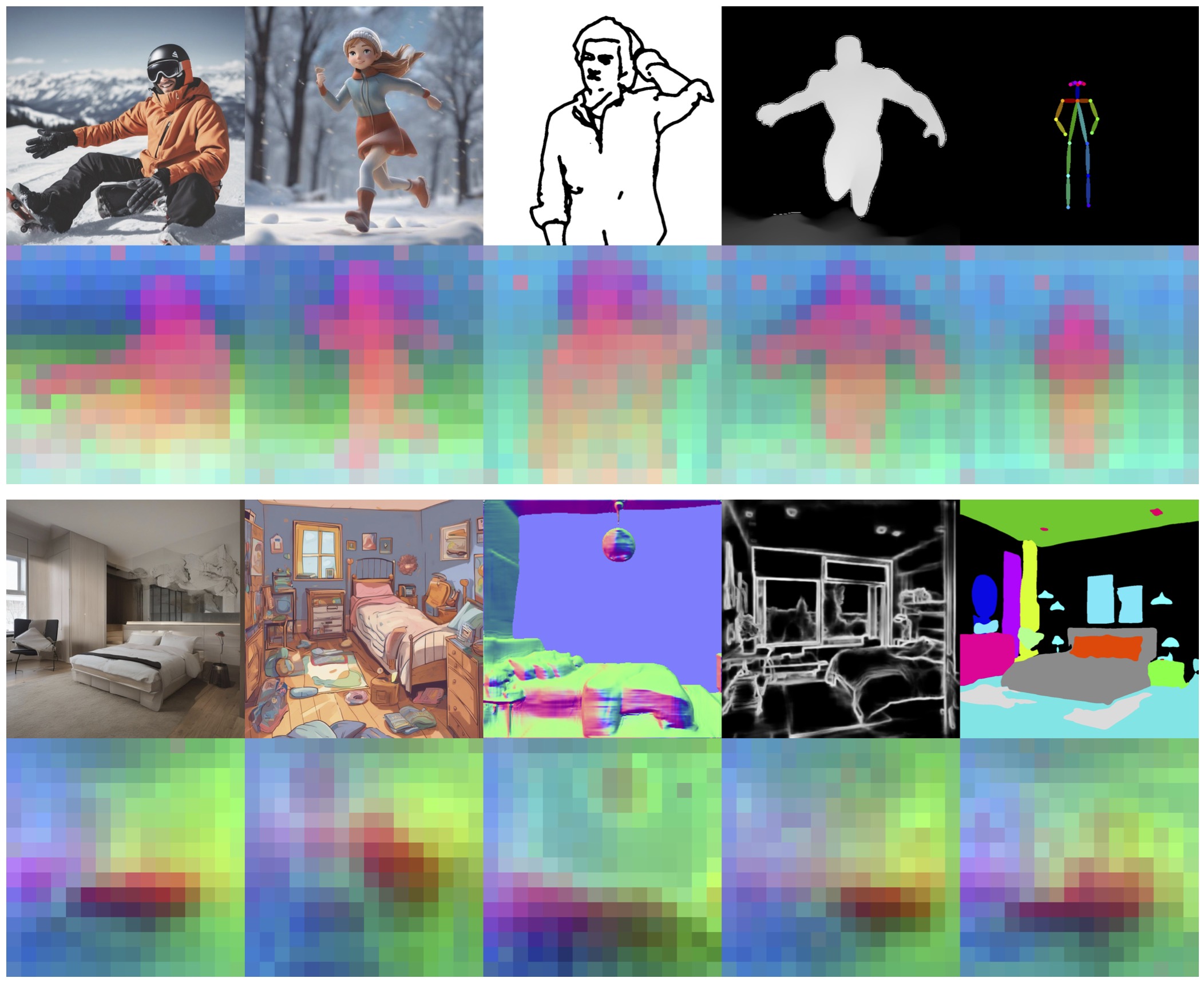}
\caption{\textbf{Visualization of feature subspace given by PCA.}
Keys from the first self-attention in the U-Net decoder are obtained via DDIM inversion~\cite{song2020ddim} for five images in different styles and modalities (\emph{top}: \texttt{person}; \emph{bottom}: \texttt{bedroom}), and subsequently undergo PCA. The top three principal components (pseudo-colored in RGB) provide a clear separation of semantic components.
}
\label{fig:pca}
\end{figure}

\noindent\textbf{Diffusion sampling.} Image generation with a pre-trained T2I diffusion model amounts to iteratively removing noise from an initial Gaussian noise image $\mathbf{x}_T$~\cite{ho2020ddpm}. This sampling process is governed by a learned denoising network $\epsilon_{\theta}$ conditioned on a text prompt $\mathbf{c}$. At a sampling step $t$, a cleaner image $\mathbf{x}_{t-1}$ is obtained by subtracting from $\mathbf{x}_t$ a noise component $\epsilon_t=\epsilon_{\theta}(\mathbf{x}_t;t,\mathbf{c})$. Alternatively, $\epsilon_{\theta}$ can be seen as approximating the score function for the marginal distributions $p_t$ scaled by a noise schedule $\sigma_t$~\cite{song2020score}:
\begin{equation}\label{eq:score}
    \epsilon_{\theta}(\mathbf{x}_t;t,\mathbf{c}) \approx -\sigma_t\nabla_{\mathbf{x}_t}\log p_t(\mathbf{x_t|\mathbf{c}}).
\end{equation}

\smallskip
\noindent\textbf{Guidance.}
The update rule in Equation~\ref{eq:score} may be altered by a time-dependent energy function $g(\mathbf{x_t};t,y)$ through \emph{guidance} (with strength $s$)~\cite{dhariwal2021classifier,epstein2023selfguidance} so as to condition diffusion sampling on auxiliary information $y$ (\eg, class labels):
\begin{equation}\label{eq:guidance}
    \hat{\epsilon}_{\theta}(\mathbf{x}_t;t,\mathbf{c}) = \epsilon_{\theta}(\mathbf{x}_t;t,\mathbf{c}) - s\,g(\mathbf{x_t};t,y).
\end{equation}

In practice, $g$ may be realized as classifiers~\cite{dhariwal2021classifier} or CLIP scores~\cite{nichol2021glide}, or defined using bounding boxes~\cite{xie2023boxdiff,chen2023layout}, attention maps~\cite{ge2023expressive,parmar2023pix2pix-zero} or any measurable object properties~\cite{epstein2023selfguidance}.

\smallskip
\noindent\textbf{Attentions in $\epsilon_{\theta}$.}
A standard choice for $\epsilon_{\theta}$ is a U-Net~\cite{ronneberger2015unet} with self- and cross-attentions~\cite{vaswani2017attention} at multiple resolutions. Conceptually, self-attentions model interactions among spatial locations within an image, whereas cross-attentions relate spatial locations to tokens in a text prompt. These two attention mechanisms complement one another and jointly control the layout of a generated image~\cite{tumanyan2023plug-and-play,cao2023masactrl,patashnik2023localizing,ge2023expressive}.

\begin{figure*}
\centering
\includegraphics[width=\linewidth]{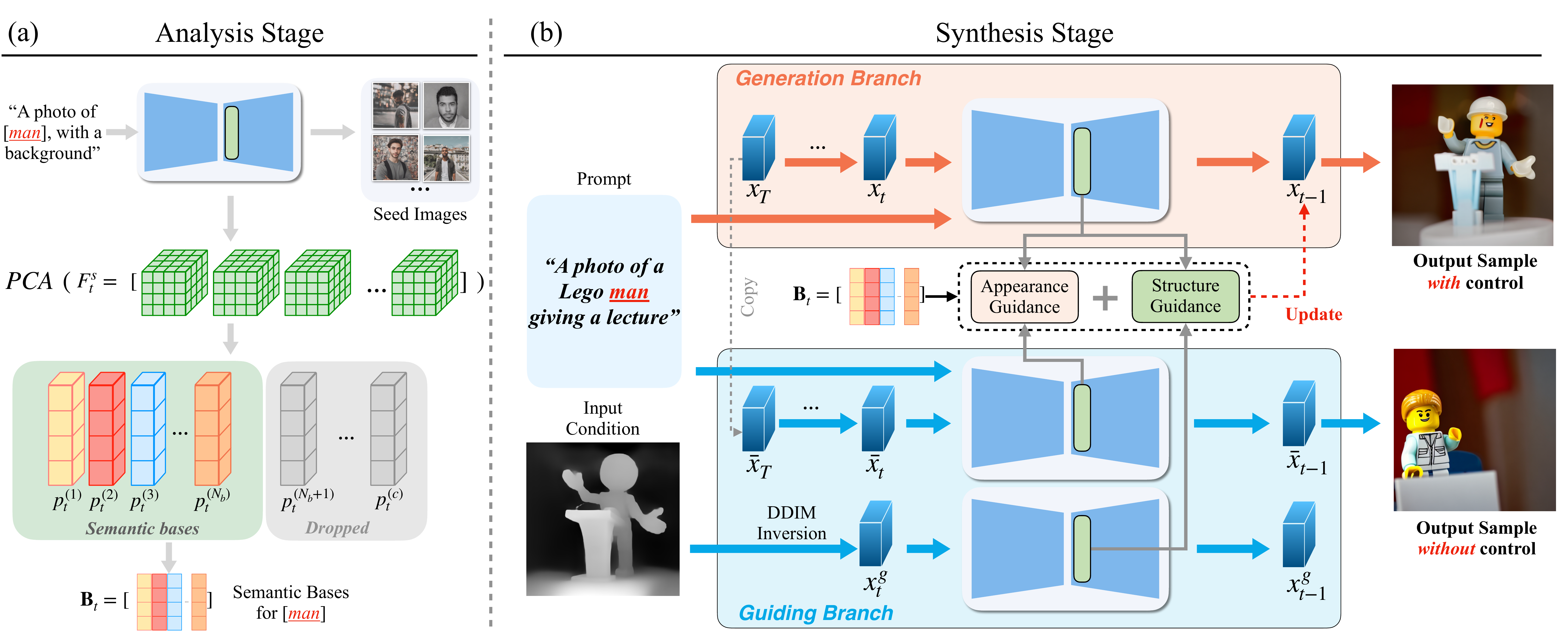}
\caption{\textbf{Method overview.}
(a) In the \emph{analysis} stage, FreeControl generates seed images for a target concept (\eg, \texttt{man}) using a pretrained diffusion model and performs PCA on their diffusion features to obtain a linear subspace as semantic basis. (b) In the \emph{synthesis} stage, FreeControl employs structure guidance in this subspace to enforce structure alignment with the input condition. In the meantime, it applies appearance guidance to facilitate appearance transfer from a sibling image generated using the same seed without structure control.
}
\label{fig:pipeline}
\end{figure*}

\section{Training-Free Control of T2I Models}
\label{sec:method}

FreeControl is a unified framework for zero-shot controllable T2I diffusion. Given a text prompt $\mathbf{c}$ and a guidance image $\mathbf{I}^g$ of any modality, FreeControl directs a pre-trained T2I diffusion model $\epsilon_{\theta}$ to comply with $\mathbf{c}$ while also respecting the semantic structure provided by $\mathbf{I}^g$ throughout the sampling process of an output image $\mathbf{I}$. 

Our key finding is that the leading principal components of self-attention block features
inside a pre-trained $\epsilon_{\theta}$ provide a strong and surprisingly consistent representation of semantic structure across a broad spectrum of image modalities (see Figure~\ref{fig:pca} for examples). To this end, we introduce \emph{structure guidance} to help draft the structural template of $\mathbf{I}$ under the guidance of $\mathbf{I}^g$. To texture this template with the content and style described by $\mathbf{c}$, we further devise \emph{appearance guidance} to borrow appearance details from $\bar{\mathbf{I}}$, a sibling of $\mathbf{I}$ generated without altering the diffusion process. Ultimately, $\mathbf{I}$ mimics the structure of $\mathbf{I}^g$ with its content and style similar to $\bar{\mathbf{I}}$.

\smallskip
\noindent\textbf{Method overview.}
FreeControl is a two-stage pipeline as illustrated in Figure~\ref{fig:pipeline}. It begins with an analysis stage, where diffusion features of \emph{seed images} undergo principal component analysis (PCA), with the leading PCs forming the time-dependent bases $\mathbf{B}_t$ as our \emph{semantic structure representation}. $\mathbf{I}^g$ subsequently undergoes DDIM inversion~\cite{song2020ddim} with its diffusion features projected onto $\mathbf{B}_t$, yielding their \emph{semantic coordinates} $\mathbf{S}^g_t$. In the synthesis stage, structure guidance encourages $\mathbf{I}$ to develop the same semantic structure as $\mathbf{I}^g$ by attracting $\mathbf{S}_t$ to $\mathbf{S}^g_t$. In the meantime, appearance guidance promotes appearance similarity between $\mathbf{I}$ and $\bar{\mathbf{I}}
$ by penalizing the difference in their feature statistics.

\subsection{Semantic Structure Representation}\label{sec:representation}

Zero-shot spatial control of T2I diffusion demands a unified representation of semantic image structure that is invariant to image modalities. Recent work has discovered that self-attention features (\ie, keys and queries) of self-supervised Vision Transformers~\cite{tumanyan2022splicing} and T2I diffusion models~\cite{cao2023masactrl} are strong descriptors of image structure. Based on these findings, we hypothesize that manipulating self-attention features is key to controllable T2I diffusion.

A na\"ive approach from PnP~\cite{tumanyan2023plug-and-play} is to directly inject the self-attention weights (equivalently the features) of $\mathbf{I}^g$ into the diffusion process of $\mathbf{I}$. Unfortunately, this approach introduces \emph{appearance leakage}; that is, not only the structure of $\mathbf{I}^g$ is carried over but also traces of appearance details. As seen in Figure~\ref{fig:cond2img_baseline}, appearance leakage is particularly problematic when $\mathbf{I}^g$ and $\mathbf{I}$ are different modalities (\eg, depth \vs natural images), common for controllable generation.

Towards disentangling image structure and appearance, we draw inspiration from Transformer feature visualization~\cite{oquab2023dinov2,tumanyan2023plug-and-play} and perform PCA on self-attention features of semantically similar images. Our key observation is that the leading PCs form a \emph{semantic basis}; It exhibits a strong correlation with object pose, shape, and scene composition across diverse image modalities. In the following, we leverage this basis as our \emph{semantic structure representation} and explain how to obtain such bases in the analysis stage.

\subsection{Analysis Stage}\label{sec:analysis}

\noindent\textbf{Seed images.}
We begin by collecting $N_s$ images that share the target concept with $\textbf{c}$. These \emph{seed images} $\{\mathbf{I}^s\}$ are generated with $\epsilon_{\theta}$ using a text prompt $\tilde{\textbf{c}}$ modified from $\textbf{c}$. Specifically, $\tilde{\textbf{c}}$ inserts the concept tokens into a template that is intentionally kept generic (\eg, ``\texttt{A photo of [] with background.}"). Importantly, this allows $\{\mathbf{I}^s\}$ to cover diverse object shape, pose, and appearance as well as image composition and style, which is key to the expressiveness of \emph{semantic bases}. We study the choice of $N_s$ in Section~\ref{sec:ablation}.

\smallskip
\noindent\textbf{Semantic basis.}
We apply DDIM inversion~\cite{song2020ddim} on $\{\mathbf{I}^s\}$ to obtain time-dependent diffusion features $\{\mathbf{F}^s_t\}$ of size ${N_s\times C \times H \times W}$ from $\epsilon_{\theta}$. This yields $N_s\times H\times W$ distinct feature vectors, on which we perform PCA to obtain the time-dependent semantic bases $\mathbf{B}_t$ as the first $N_b$ principal components:
\begin{equation}
    \mathbf{B}_t=[\mathbf{p}^{(1)}_t,\mathbf{p}^{(2)}_t,...,\mathbf{p}^{(N_b)}_t]\sim\mathrm{PCA}(\{\mathbf{F}^s_t\})
\end{equation}

Intuitively, $\mathbf{B}_t$ span semantic spaces $\mathbb{S}_t$ that connect different image modalities, allowing the propagation of image structure from $\mathbf{I}^g$ to $\mathbf{I}$ in the synthesis stage. We study the choice of $\mathbf{F}_t$ and $N_b$ in Section~\ref{sec:ablation}.

\smallskip
\noindent\textbf{Basis reuse.} Once computed, $\mathbf{B}_t$ can be reused for the same text prompt or shared by prompts with related concepts. The cost of basis construction can thus be amortized over multiple runs of the synthesis stage.

\begin{figure*}
\centering
\includegraphics[width=\linewidth]{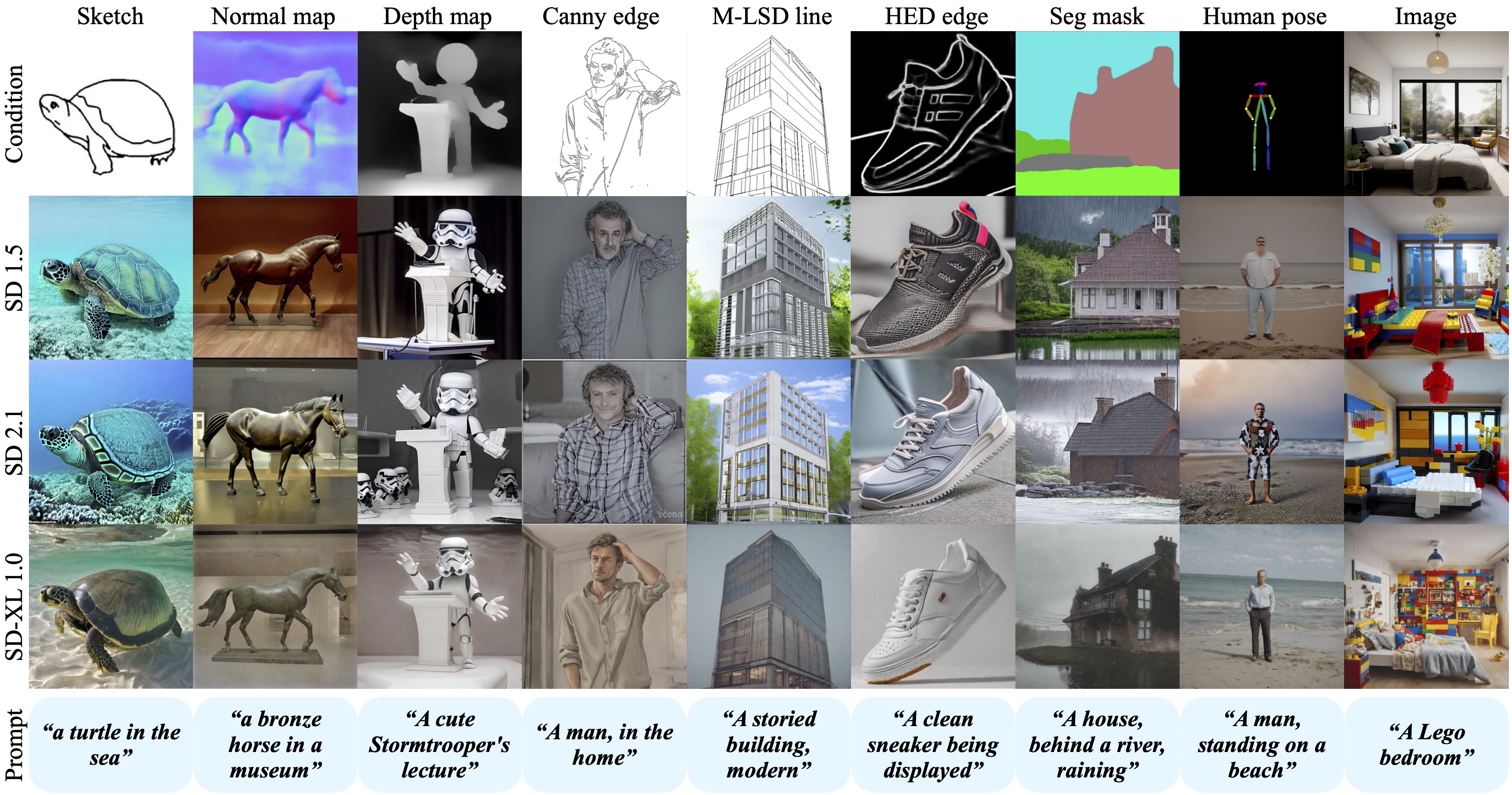}
\caption{\textbf{Qualitative comparison of controllable T2I diffusion.} FreeControl supports a suite of control signals and three major versions of Stable Diffusion. The generated images closely follow the text prompts while exhibiting strong spatial alignment with the input images.
}
\label{fig:cond2img_qualitative}
\end{figure*}

\subsection{Synthesis Stage}\label{sec:synthesis}

The generation of $\mathbf{I}$ is conditioned on $\mathbf{I}^g$ through guidance. As a first step, we express the semantic structure of $\mathbf{I}^g$ with respect to the semantic bases $\mathbf{B}_t$.

\smallskip
\noindent\textbf{Inversion of $\mathbf{I}^g$.}
We perform DDIM inversion~\cite{song2020ddim} on $\mathbf{I}^g$ to obtain the diffusion features $\mathbf{F}^g_t$ of size $C\times H\times W$ and project them onto $\mathbf{B}_t$ to obtain their \emph{semantic coordinates} $\mathbf{S}^g_t$ of size $N_b\times H\times W$. For local control of foreground structure, we further derive a mask $\mathbf{M}$ (size $H\times W$) from cross-attention maps of the concept tokens~\cite{ge2023expressive}. $\mathbf{M}$ is set to $\mathbf{1}$ (size ${H\times W}$) for global control.

\smallskip
We are now ready to generate $\mathbf{I}$ with \emph{structure guidance} to control its underlying semantic structure.

\smallskip
\noindent\textbf{Structure guidance.}
At each denoising step $t$, we obtain the semantic coordinates $\mathbf{S}_t$ by projecting the diffusion features $\mathbf{F}_t$ from $\epsilon_{\theta}$ onto $\mathbf{B}_t$. Our energy function $g_s$ for structure guidance can then be expressed as
\begin{equation*}
\small
\begin{split}
    g_s(\mathbf{S}_t;\mathbf{S}^g_t,\mathbf{M}) & =  \underbrace{\frac{\sum_{i,j}m_{ij}\|[\mathbf{s}_t]_{ij}-[\mathbf{s}^g_t]_{ij}\|^2_2}{\sum_{i,j} m_{ij}}}_\text{forward guidance} \\ & + w\cdot \underbrace{\frac{\sum_{i,j}(1-m_{ij})\|\max([\mathbf{s}_t]_{ij}-\boldsymbol{\tau}_t, 0)\|^2_2}{\sum_{i,j}(1-m_{ij})}}_\text{backward guidance},
\end{split}
\end{equation*}
where $i$ and $j$ are spatial indices for $\mathbf{S}_t$, $\mathbf{S}^g_t$ and $\mathbf{M}$, and $w$ is the balancing weight. The thresholds $\boldsymbol{\tau}_t$ are defined as
\begin{equation}
    \boldsymbol{\tau}_t=\max_{i,j\:\mathrm{s.t.}\: m_{ij}=0}\:[\mathbf{s}^g_t]_{ij}
\end{equation}
with $\max$ taken per channel. Loosely speaking, $[\mathbf{s}_t]_{ij}>\boldsymbol{\tau}_t$ indicates the presence of foreground structure. Intuitively, the \emph{forward} term guides the structure of $\mathbf{I}$ to align with $\mathbf{I}^g$ in the foreground, whereas the \emph{backward} term, effective when $\mathbf{M}\not=\mathbf{1}$, helps carve out the foreground by suppressing spurious structure in the background.

\smallskip
While structure guidance drives $\mathbf{I}$ to form the same semantic structure as $\mathbf{I}^g$, we found that it also amplifies low-frequency textures, producing cartoony images that lack appearance details. To fix this problem, we apply \emph{appearance guidance} to borrow texture from $\bar{\mathbf{I}}$, a sibling image of $\mathbf{I}$ generated from the same noisy latent with the same seed yet without structure guidance.

\smallskip
\noindent\textbf{Appearance representation.} Inspired by DSG~\cite{epstein2023selfguidance}, we represent image appearance as $\{\mathbf{v}^{(k)}_t\}^{N_a\leq N_b}_{k=1}$, the weighted spatial means of diffusion features $\mathbf{F}_t$:
\begin{equation}
    \mathbf{v}^{(k)}_t=\frac{\sum_{i,j}\sigma([s^{(k)}_t]_{ij})[\mathbf{f}_t]_{ij}}{\sum_{i,j}\sigma([s^{(k)}_t]_{ij})},
\end{equation}
where $i$ and $j$ are spatial indices for $\mathbf{S}_t$ and $\mathbf{F}_t$, $k$ is channel index for $[\mathbf{s}_t]_{i,j}$, and $\sigma$ is the sigmoid function. We repurpose $\mathbf{S}_t$ as weights so that different $\mathbf{v}^{(k)}_t$'s encode appearance of distinct semantic components. We calculate $\{\mathbf{v}^{(k)}_t\}$ and $\{\bar{\mathbf{v}}^{(k)}_t\}$ respectively for $\mathbf{I}$ and $\bar{\mathbf{I}}$ at each timestep $t$.

\smallskip
\noindent\textbf{Appearance guidance.}
Our energy function $g_a$ for appearance guidance can then be expressed as
\begin{equation}
    g_a(\{\mathbf{v}^{(k)}_t\};\{\bar{\mathbf{v}}^{(k)}_t\})=\frac{\sum^{N_a}_{k=1}\|\mathbf{v}^{(k)}_t-\bar{\mathbf{v}}^{(k)}_t\|^2_2}{N_a}.
\end{equation}
It penalizes difference in the appearance representations and thus facilitates appearance transfer from $\bar{\mathbf{I}}$ to $\mathbf{I}$.

\smallskip
\noindent\textbf{Guiding the generation process.}
Finally, we arrive at our modified score estimate $\hat{\epsilon}_t$ by including structure and appearance guidance alongside classifier-free guidance~\cite{ho2022cfg}:
\begin{equation}
    \hat{\epsilon}_t=(1+s)\,\epsilon_{\theta}(\mathbf{x}_t;t,\mathbf{c})-s\,\epsilon_{\theta}(\mathbf{x}_t;t,\emptyset)+ \lambda_s\,g_s + \lambda_a\,g_a
\end{equation}
where $s$, $\lambda_s$ and $\lambda_a$ are the respective guidance strengths.
\section{Experiments and Results}
\label{sec:experiments}

We report extensive qualitative and quantitative results to demonstrate the effectiveness and generality of our approach for zero-shot controllable T2I diffusion. We present additional results on text-guided image-to-image translation and provide ablation studies on key model components.

\begin{figure*}[htp]
\centering
\includegraphics[width=0.925\linewidth]{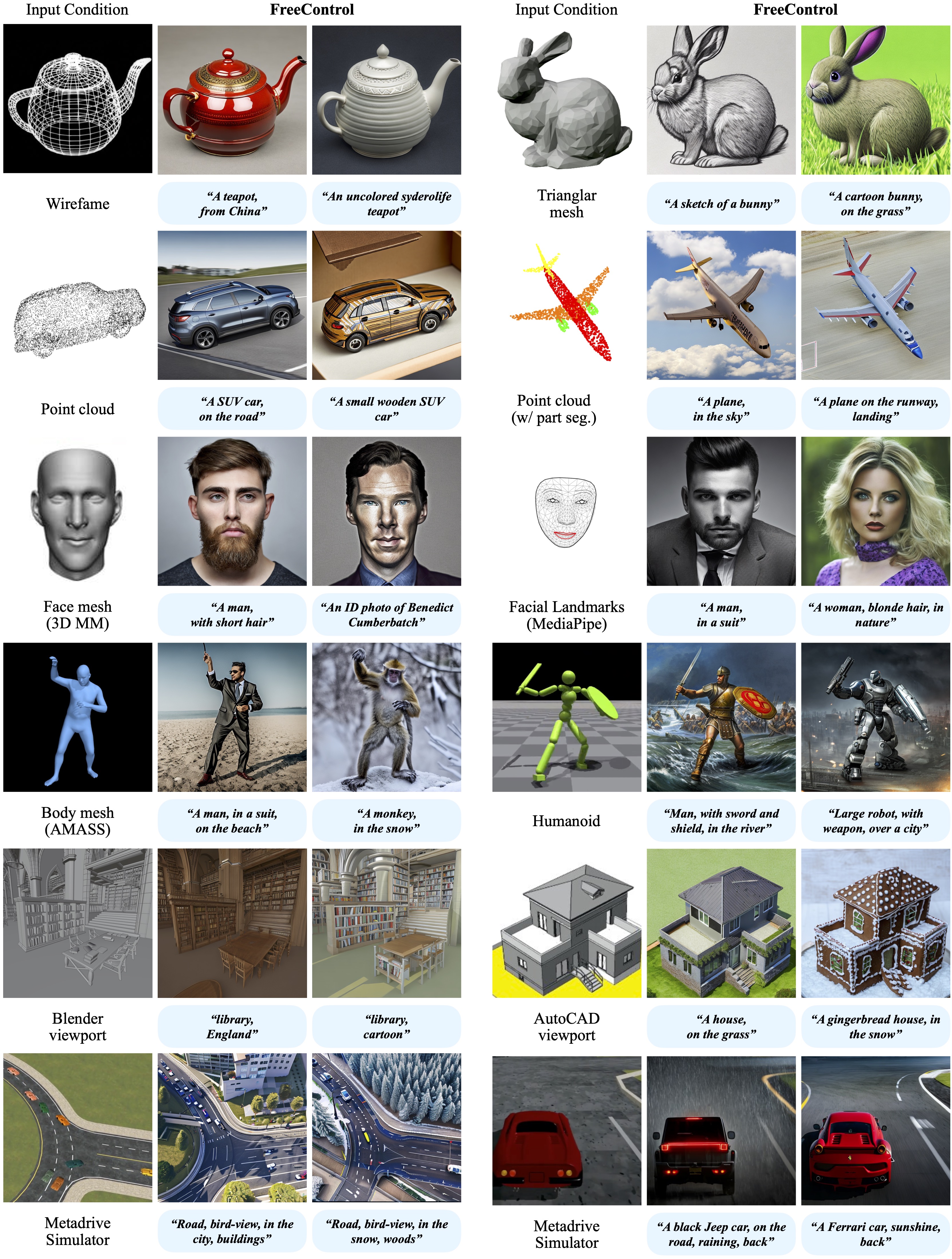}
\caption{\textbf{Qualitative results for more control conditions.} FreeControl supports challenging control conditions not possible with training-based methods. These include 2D projections of common graphics primitives \emph{(row 1 and 2)}, domain-specific shape models \emph{(row 3 and 4)}, graphics software viewports \emph{(row 5)}, and simulated driving environments \emph{(row 6)}.}
\label{fig:any_cond}
\end{figure*}

\subsection{Controllable T2I Diffusion}

\smallskip
\noindent\textbf{Baselines.}
ControlNet~\cite{zhang2023controlNet} and T2I-Adapter~\cite{mou2023t2i} learn an auxiliary module to condition a pretrained diffusion model on a guidance image. One such module is learned for each condition type. Uni-ControlNet~\cite{zhao2023uni} instead learns adapters shared by all condition types for all-in-one control. Different from these training-based methods, SDEdit~\cite{meng2021sdedit} adds noise to a guidance image and subsequently denoises it with a pretrained diffusion model for guided image synthesis. Prompt-to-Prompt (P2P)~\cite{hertz2022prompt-to-prompt} and Plug-and-Play (PnP)~\cite{tumanyan2023plug-and-play} manipulate attention weights and features inside pretrained diffusion models for zero-shot image editing. We compare our method with these strong baselines in our experiments.

\smallskip
\noindent\textbf{Experiment setup.}
Similar to ControlNet~\cite{zhang2023controlNet}, we report qualitative results on eight condition types (sketch, normal, depth, Canny edge, M-LSD line, HED edge, segmentation mask, and human pose). We further employ several previously unseen control signals as input conditions (Figure~\ref{fig:any_cond}), and combine our method with all major versions of Stable Diffusion (1.5, 2.1, and XL 1.0) to study its generalization on diffusion model architectures.

For a fair comparison with the baselines, we adapt the ImageNet-R-TI2I dataset from PnP~\cite{tumanyan2023plug-and-play} as our benchmark dataset. It contains 30 images from 10 object categories. Each image is associated with five text prompts originally for the evaluation of text-guided image-to-image translation. We convert the images into their respective Canny edge, HED edge, sketch, depth map, and normal map following ControlNet~\cite{zhang2023controlNet}, and subsequently use them as input conditions for all methods in our experiments.

\smallskip
\noindent\textbf{Evaluation metrics.}
We report three widely adopted metrics for quantitative evaluation; \emph{Self-similarity distance}~\cite{tumanyan2022splicing} measures the structural similarity of two images in the feature space of DINO-ViT~\cite{caron2021dino}. A smaller distance suggests better structure preservation. Similar to~\cite{tumanyan2023plug-and-play}, we report self-similarity between the generated image and the dataset image that produces the input condition. \emph{CLIP score}~\cite{radford2021clip} measures image-text alignment in the CLIP embedding space. A higher CLIP score indicates a stronger semantic match between the text prompt and the generated image. \emph{LPIPS distance}~\cite{zhang2018lpips} measures the appearance deviation of the generated image from the input condition. Images with richer appearance details yield higher LPIPS score.

\smallskip
\noindent\textbf{Implementation details.}
We adopt keys from the first self-attention in the U-Net decoder as the features $\mathbf{F}_t$. We run DDIM inversion on $N_s=20$ seed images for $200$ steps to obtain bases of size $N_b=64$. In the synthesis stage, we run DDIM inversion on $\mathbf{I}^g$ for $1000$ steps, and sample $\mathbf{I}$ and $\bar{\mathbf{I}}$ by running $200$ steps of DDIM sampling. Structure and appearance guidance are applied in the first $120$ steps. $\lambda_s\in[400, 1000]$, $\lambda_a=0.2\lambda_s$, and $N_a=2$ in all experiments.

\begin{figure*}
\centering
\includegraphics[width=\linewidth]{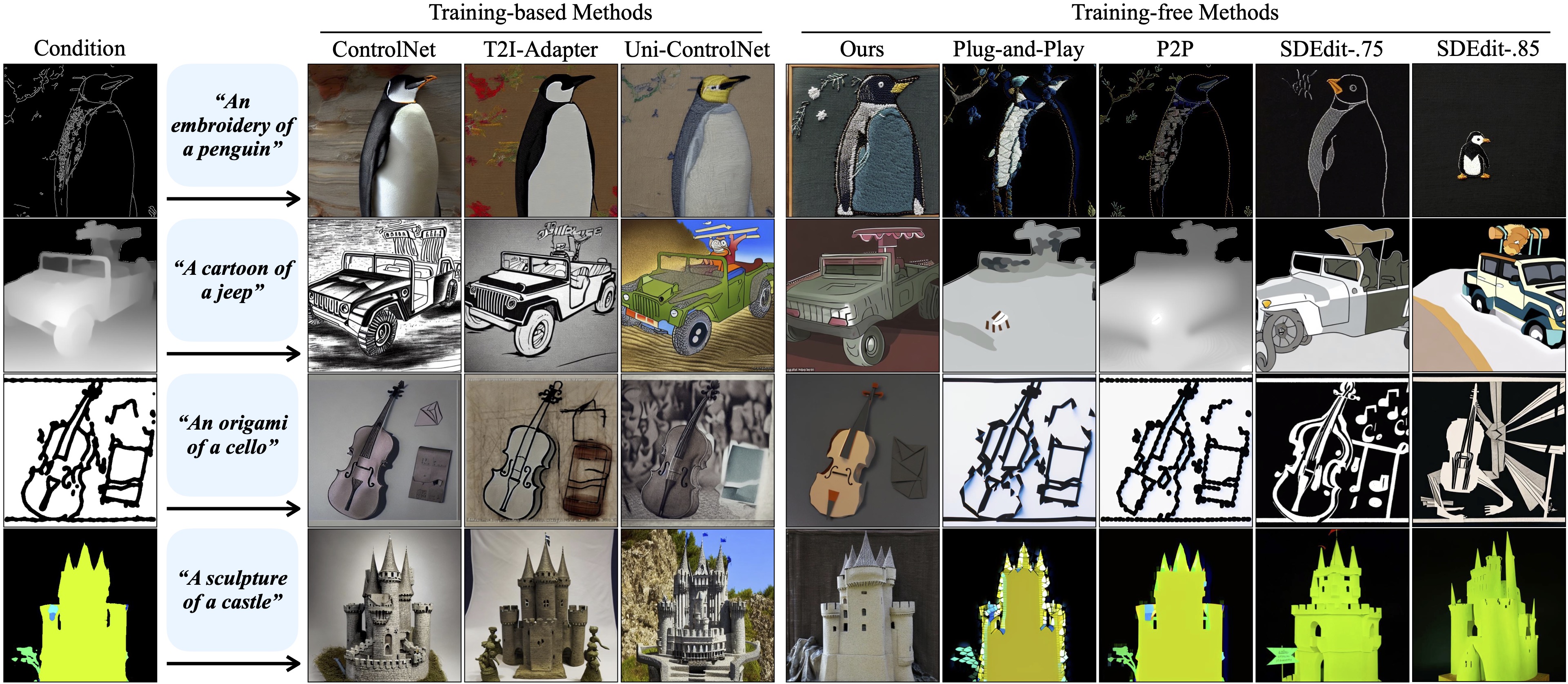}
\caption{\textbf{Qualitative comparison on controllable T2I diffusion.}
FreeControl achieves competitive spatial control and superior image-text alignment in comparison to training-based methods. It also escapes the appearance leakage problem manifested by the training-free baselines, producing high-quality images with rich content and appearance faithful to the text prompt.
}
\label{fig:cond2img_baseline}
\end{figure*}

\begin{table*}[]
\resizebox{\linewidth}{16.5mm}{
\begin{tabular}{lccccccccccccccc}
\toprule[1.5pt]
\multirow{2}{*}{Method} & \multicolumn{3}{c}{Canny} & \multicolumn{3}{c}{HED} & \multicolumn{3}{c}{Sketch} & \multicolumn{3}{c}{Depth} & \multicolumn{3}{c}{Normal} \\ \cmidrule(r){2-4} \cmidrule(r){5-7} \cmidrule(r){8-10} \cmidrule(r){11-13} \cmidrule(r){14-16}
               & Self-Sim~$\downarrow$ & CLIP~$\uparrow$  & LPIPS~$\uparrow$ & Self-Sim~$\downarrow$ & CLIP~$\uparrow$  & LPIPS~$\uparrow$ & Self-Sim~$\downarrow$ & CLIP~$\uparrow$  & LPIPS~$\uparrow$ & Self-Sim~$\downarrow$ & CLIP~$\uparrow$  & LPIPS~$\uparrow$ & Self-Sim~$\downarrow$ & CLIP~$\uparrow$  & LPIPS~$\uparrow$ \\ \midrule[0.5pt]
\color[HTML]{9b9b9b}ControlNet~\cite{zhang2023controlNet}     & \color[HTML]{9b9b9b}0.042    & \color[HTML]{9b9b9b}0.300 & \color[HTML]{9b9b9b}0.665 &  \color[HTML]{9b9b9b}0.040    & \color[HTML]{9b9b9b}0.291 & \color[HTML]{9b9b9b}0.609 & \color[HTML]{9b9b9b}0.070    & \color[HTML]{9b9b9b}0.314 & \color[HTML]{9b9b9b}0.668 &  \color[HTML]{9b9b9b}0.058    & \color[HTML]{9b9b9b}0.306 & \color[HTML]{9b9b9b}0.645 & \color[HTML]{9b9b9b}0.079    & \color[HTML]{9b9b9b}0.304 & \color[HTML]{9b9b9b}0.637 \\
\color[HTML]{9b9b9b}T2I-Adapter    & \color[HTML]{9b9b9b}0.052    & \color[HTML]{9b9b9b}0.290 & \color[HTML]{9b9b9b}0.689 & \color[HTML]{9b9b9b}-        & \color[HTML]{9b9b9b}-     & \color[HTML]{9b9b9b}-     & \color[HTML]{9b9b9b}0.096    & \color[HTML]{9b9b9b}0.290 & \color[HTML]{9b9b9b}0.648 \color[HTML]{9b9b9b}& \color[HTML]{9b9b9b}0.071    & \color[HTML]{9b9b9b}0.314 & \color[HTML]{9b9b9b}0.673 & \color[HTML]{9b9b9b}-        & \color[HTML]{9b9b9b}-     & \color[HTML]{9b9b9b}-      \\
\color[HTML]{9b9b9b}Uni-ControlNet & \color[HTML]{9b9b9b}0.044    & \color[HTML]{9b9b9b}0.295 & \color[HTML]{9b9b9b}0.539 & \color[HTML]{9b9b9b}0.050    & \color[HTML]{9b9b9b}0.301 & \color[HTML]{9b9b9b}0.553 & \color[HTML]{9b9b9b}0.050    & \color[HTML]{9b9b9b}0.301 & \color[HTML]{9b9b9b}0.553 & \color[HTML]{9b9b9b}0.061    & \color[HTML]{9b9b9b}0.303 & \color[HTML]{9b9b9b}0.636 & \color[HTML]{9b9b9b}-        &  \color[HTML]{9b9b9b}-    & \color[HTML]{9b9b9b}-   \\ 
\addlinespace
SDEdit-0.75~\cite{meng2021sdedit}    & 0.108    & 0.306 & 0.582 & 0.123    & 0.288 & 0.375 & 0.135    & 0.281 & 0.361 & 0.153    & 0.294 & 0.327 & 0.128    & 0.284 & 0.456 \\
SDEdit-0.85~\cite{meng2021sdedit}    & 0.139    & 0.319 & \textbf{0.670} & 0.153    & 0.305 & 0.485 & 0.139    & 0.300 & 0.485 & 0.165    & 0.304 & 0.384 & 0.147    & 0.298 & 0.512 \\
P2P~\cite{hertz2022prompt-to-prompt}            & 0.078    & 0.253 & 0.298 & 0.112    & 0.253 & 0.194 & 0.194    & 0.251 & 0.096 & 0.142    & 0.248 & 0.167 & 0.100    & 0.249 & 0.198 \\
PNP~\cite{tumanyan2023plug-and-play} & \textbf{0.074}    & 0.282 & 0.417 & 0.098    & 0.286 & 0.271 & 0.158    & 0.267 & 0.221 & 0.126    & 0.287 & 0.268 & 0.107    & 0.286 & 0.347  \\
\rowcolor{gray!10} Ours           & \textbf{0.074}    & \textbf{0.338} & 0.667 & \textbf{0.075}  &  \textbf{0.337} & \textbf{0.561} & \textbf{0.086}    & \textbf{0.337} & \textbf{0.593} & \textbf{0.077}    & \textbf{0.307} & \textbf{0.477} & \textbf{0.086}    & \textbf{0.335} & \textbf{0.629} \\ 
 \bottomrule[1.5pt]
\end{tabular}
}
\centering
\caption{\textbf{Quantitative results on controllable T2I diffusion.}
FreeControl consistently outperforms all training-free baselines in structure preservation, image-text alignment and appearance diversity as measured by Self-similarity distance, CLIP score and LPIPS distance. It achieves competitive structure and appearance scores with the training-based baselines while demonstrate stronger image-text alignment.
}\label{tab:cond2img}
\end{table*}

\smallskip
\noindent\textbf{Qualitative results.}
As shown in Figure~\ref{fig:cond2img_qualitative}, FreeControl is able to recognize diverse semantic structures from all condition modalities used by ControlNet~\cite{zhang2023controlNet}. It produces high-quality images in close alignment with both the text prompts and spatial conditions. Importantly, it generalizes well on all major versions of Stable Diffusion, enabling effortless upgrade to future model architectures without retraining.

In Figure~\ref{fig:any_cond}, we present additional results for condition types not possible with previous methods. FreeControl generalizes well across challenging condition types for which constructing training pairs is difficult. In particular, it enables superior conditional control with common graphics primitives (\eg, mesh and point cloud), domain-specific shape models (\eg, face and body meshes), graphics software viewports (\eg, Blender~\cite{blender} and AutoCAD~\cite{autocad}), and simulated driving environments (\eg, MetaDrive~\cite{li2021metadrive}), thereby providing an appealing solution to visual design preview and sim2real.

\begin{figure*}[h]
\centering
\includegraphics[width=\linewidth]{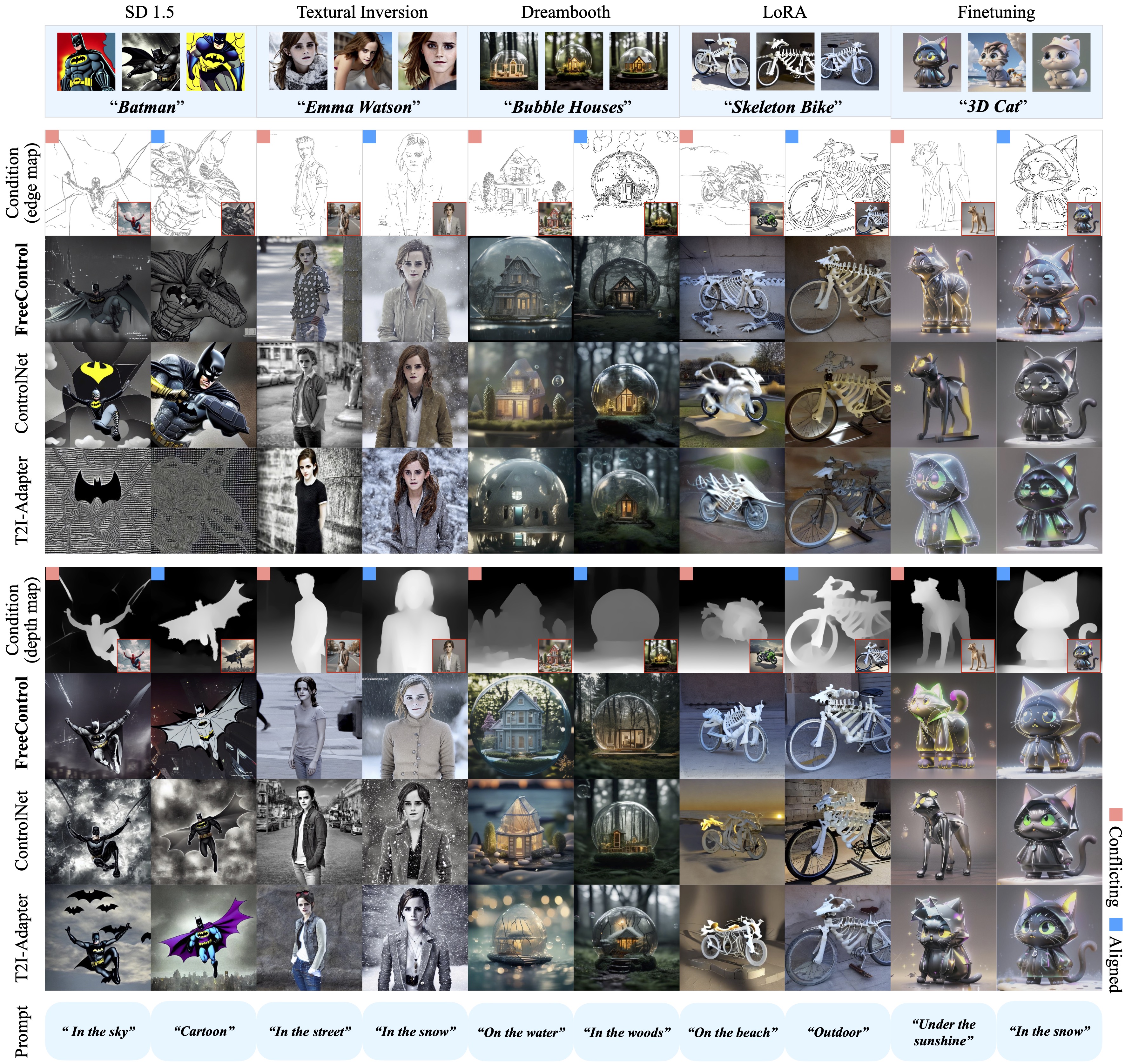}
\caption{\textbf{Controllable T2I generation of custom concepts.}
FreeControl is compatible with major customization techniques and readily supports controllable generation of custom concepts without requiring spatially-aligned condition images. By contrast, ControlNet fails to preserve custom concepts given conflicting conditions, whereas T2I-Adapter refuses to respect the condition image and text prompt.
}
\label{fig:cond2img_custom}
\end{figure*}

\smallskip
\noindent\textbf{Comparison with baselines.}
Figure~\ref{fig:cond2img_baseline} and Table~\ref{tab:cond2img} compare our methods to the baselines. Despite stronger structure preservation (\ie, small self-similarity distances), the training-based methods at times struggle to follow the text prompt (\eg \textit{embroidery} for ControlNet and \textit{origami} for all baselines) and yield worse CLIP scores. The loss of text control is a common issue in training-based methods due to modifications made to the pretrained models. Our method is training-free, hence retaining strong text conditioning.

On the other hand, training-free baselines are prone to appearance leakage as a generated image shares latent states (SDEdit) or diffusion features (P2P and PnP) with the condition image. As a result, not only is the structure carried over but also undesired appearance, resulting in worse LPIPS scores. For example, all baselines inherit the texture-less background in the \textit{embroidery} example and the foreground shading in the \textit{castle} example. By contrast, our method decouples structure and appearance, thereby escaping appearance leakage.

\smallskip
\noindent\textbf{Handling conflicting conditions.} Finally, we study cases where spatial conditions have minor conflicts to input text prompts. We assume that a text prompt consists of a concept (\eg, batman) and a style (\eg, cartoon), and contrast a conflicting case with its aligned version. Specifically, a conflicting case includes (a) a text prompt with a feasible combination of concept and style; and (b) a spatial condition (\ie an edge map) derived from real images without the text concept. The corresponding aligned case contains a similar text prompt, yet using a spatial condition from real images with the same concept. We input those cases into ControlNet, T2I-Adapter, and  FreeControl, using a set of pre-trained and customized models. 

Figure~\ref{fig:cond2img_custom} shows the results. Our training-free FreeControl consistently generates high quality images that fit the middle ground of spatial conditions and text prompts, across all test cases and models. T2I-Adapter sometimes fails even with an aligned case (see \textit{Batman} examples), not to mention the conflicting cases. Indeed, T2I-Adapter tends to disregard the condition image, leading to diminished controllability, as exemplified by \textit{Emma Watson} example (conflicting). ControlNet can generate convincing images for aligned cases, yet often fall short in those conflicting cases. A common failure mode is to overwrite the input text concept using the condition image, as shown by \textit{skeleton bike} or \textit{house in a bubble} examples (conflicting).

\begin{figure*}
    \centering
    \includegraphics[width=\linewidth]{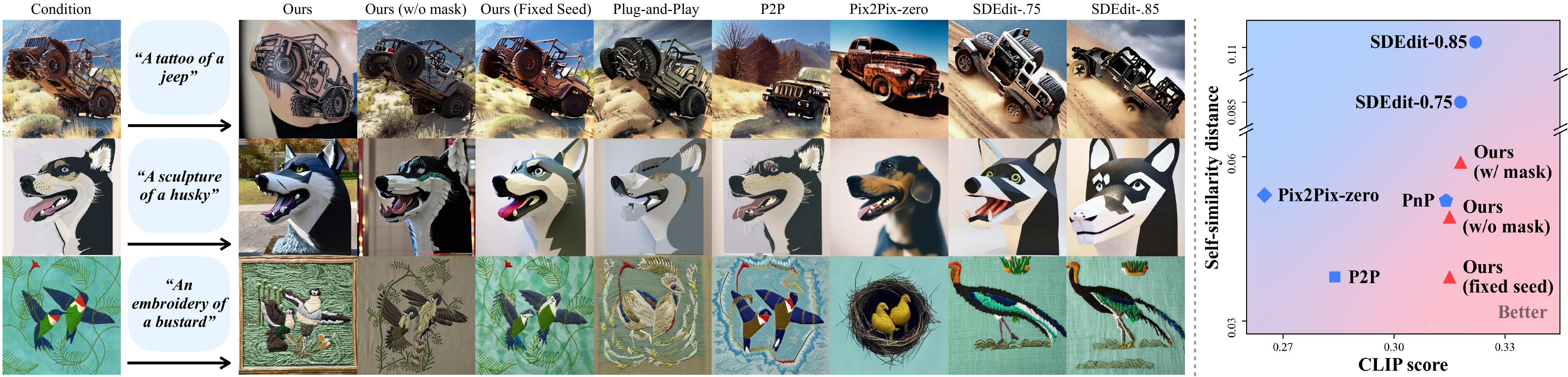}
    \caption{\textbf{Qualitative and quantitative comparison on text-guided image-to-image translation.}
    FreeControl enables flexible control of image composition and style through guidance mask $\mathbf{M}$ and random seed (\emph{left}). It strikes a good balance between structure preservation (self-similarity distance) and image-text alignment (CLIP score) in comparison to the baselines (\emph{right}, better towards bottom right).
    }
    \label{fig:img2img_results}
\end{figure*}

\begin{figure*}[htp]
\centering
\includegraphics[width=\linewidth]{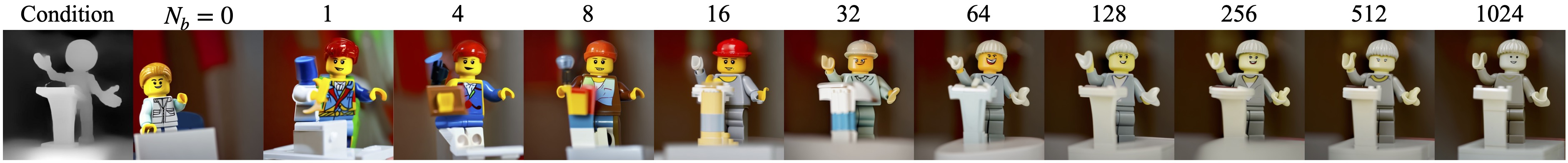}
\caption{\textbf{Ablation on size of semantic bases $N_b$.}
Images are generated using the prompt \textit{``a Lego man giving a lecture"}. They illustrate an inherent tradeoff between structure and appearance quality. A good balance can be achieved with $N_b$'s in the middle range.
}
\label{fig:ablation_on_n_b}
\end{figure*}

\subsection{Extension to Image-to-Image Translation}

FreeControl can be readily extended to support image-to-image (I2I) translation by conditioning on a detailed / real image. A key challenge here is to allow FreeControl to preserve the background provided by the condition, \ie, the input content image. To this end, we propose two variants of FreeControl. The first removes the mask $\mathbf{M}$ in structure guidance (\ie w/o mask), and the second generates from the inverted latent $\mathbf{x}^g_T$ of the condition image (\ie fixed seed). We find that removing the mask helps extract and maintain the background structure, and starting inference from $\mathbf{x}^g_T$ retains the appearance from the condition image.

Figure~\ref{fig:img2img_results} evaluates FreeControl and its two variants for text-guided I2I, and compares to strong baselines for the I2I task including PnP~\cite{tumanyan2023plug-and-play}, P2P~\cite{hertz2022prompt-to-prompt}, Pix2Pix-zero~\cite{parmar2023pix2pix-zero} and SDEdit~\cite{meng2021sdedit}. The vanilla FreeControl, as we expect, often fails to preserve the background. However, our two variants with simple modification demonstrate impressive results as compared to the baselines, generating images that adhere to both foreground and background of the input image.

Further, we evaluate the \textit{self-similarity distance} and \textit{CLIP score} of FreeControl, its variants, and our baselines on the ImageNet-R-TI2I dataset. The results are summarized in Figure~\ref{fig:img2img_results}. Variants of FreeControl outperform all baselines with significantly improved structure preservation and visual fidelity, following the input text prompts.

\subsection{Ablation Study}
\label{sec:ablation}

\begin{figure}[htp]
\centering
\includegraphics[width=\linewidth]{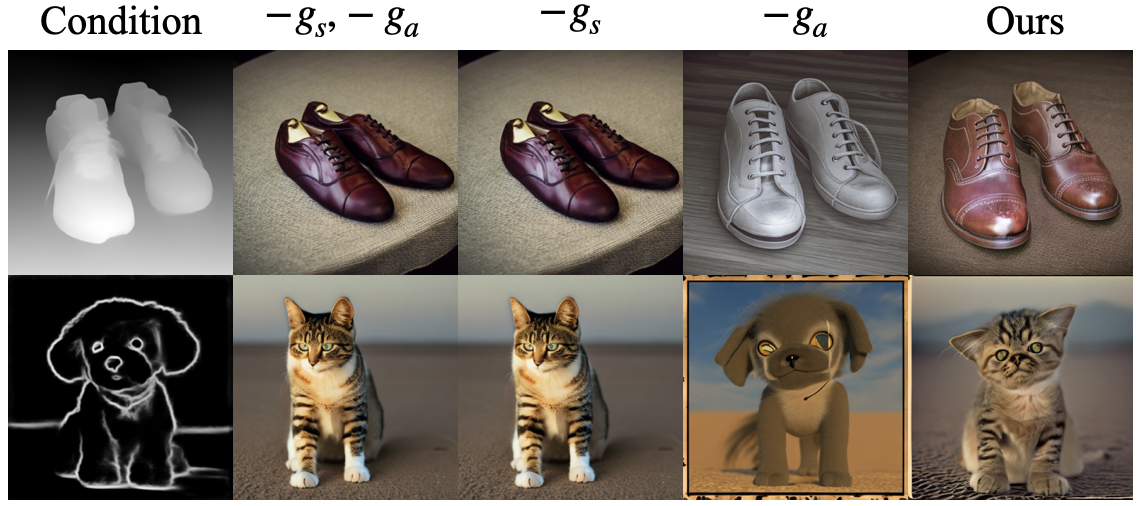}
\caption{\textbf{Ablation on guidance effect.}
Top: \textit{``leather shoes''}; Bottom: \textit{``cat, in the desert''}. $g_s$ and $g_a$ stand for structure and appearance guidance, respectively.}
\label{fig:ablation_guidance}
\end{figure}

\smallskip
\noindent\textbf{Effect of guidance.}
As seen in Figure~\ref{fig:ablation_guidance}, structure guidance is responsible for structure alignment ($-g_s$ \vs Ours). Appearance guidance alone has no impact on generation in the absence of structure guidance ($-g_a$ \vs $-g_s, -g_a$). It only becomes active after image structure has shaped up, in which case it facilitates appearance transfer ($-g_a$ \vs Ours).

\begin{figure}[htp]
\centering
\includegraphics[width=\linewidth]{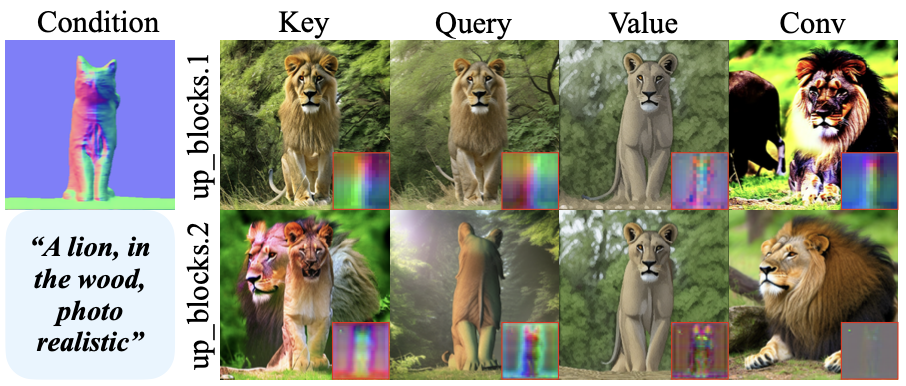}
\caption{\textbf{Ablation on feature choice.}
Keys from self-attention of up\_block.1 in the U-Net decoder expose the strongest controllability. PCA visualization of the features are in the insets.
}
\label{fig:ablation_blocks}
\end{figure}

\smallskip
\noindent\textbf{Choice of diffusion features $\mathbf{F}_t$.}
Figure~\ref{fig:ablation_blocks} compares results using self-attention keys, queries, values, and their preceding Conv features from up\_block.[1,2] in the U-Net decoder. It reveals that up\_block.1 in general carries more structural cues than up\_block.2, whereas keys better disentangle semantic components than the other features.

\begin{figure}[htp]
\centering
\includegraphics[width=\linewidth]{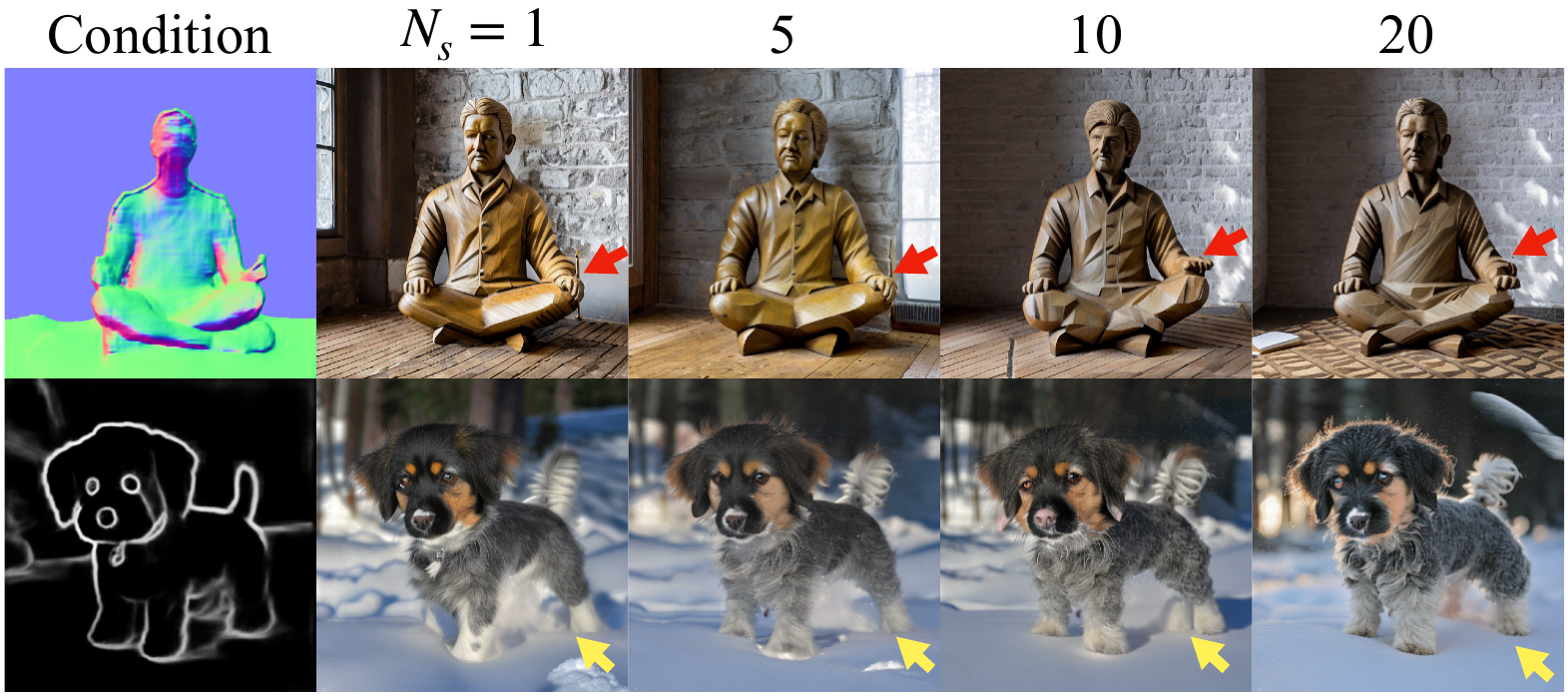}
\caption{\textbf{Ablation on number of seed images $N_s$.} Top: \textit{``wooden sculpture of a man''}; Bottom: \textit{``dog, in the snow''}. Larger $N_s$ brings minor improvement on structure alignment.}
\label{fig:ablation_on_n_s}
\end{figure}

\smallskip
\noindent\textbf{Number of seed images $N_s$.}
Figure~\ref{fig:ablation_on_n_s} suggests that $N_s$ has minor impact on image quality and controllability, allowing the use of \emph{as few as $1$ seed image} in the analysis stage. Large $N_s$ diversifies image content and style, which helps perfect structural details (\eg limbs) in the generated images.

\smallskip
\noindent\textbf{{Size of semantic bases $N_b$.}}
Figure~\ref{fig:ablation_on_n_b} presents generation results over the full spectrum of $N_b$. A larger $N_b$ improves structure alignment yet triggers the unintended transfer of appearance from the input condition. Hence, a good balance is achieved with $N_b$'s in the middle range.

\section{Conclusion}
\label{sec:conclusion}
We present FreeControl, a training-free method for spatial control of any T2I diffusion models with many conditions. FreeControl exploits the feature space of pretrained T2I models, facilitates convenient control over many architectures and checkpoints, allows various challenging input conditions on which most of the existing training-free methods fail, and achieves competitive synthesis quality with training-based approaches. One limitation is that FreeContorl relies on the DDIM inversion process to extract intermediate features of the guidance image and compute additional gradient during the synthesis stage, resulting in increased inference time. We hope our findings and analysis can shed light on controllable visual content creation.

{
    \small
    \bibliographystyle{ieeenat_fullname}
    \bibliography{main}
}

\clearpage
\section*{Supplementary Material}

\setcounter{section}{0}
\renewcommand\thesection{\Alph{section}}

In the supplementary material, we present additional qualitative results (Section~\ref{supp:qual_results}) and ablation experiments (Section~\ref{supp:ablation}), and discuss the limitations (Section~\ref{supp:limitation}) and societal impact of our method (Section~\ref{supp:impact}). We hope this document complements the main paper.

\section{Additional Qualitative Results}\label{supp:qual_results}

\begin{figure*}
\centering
\includegraphics[width=\linewidth]{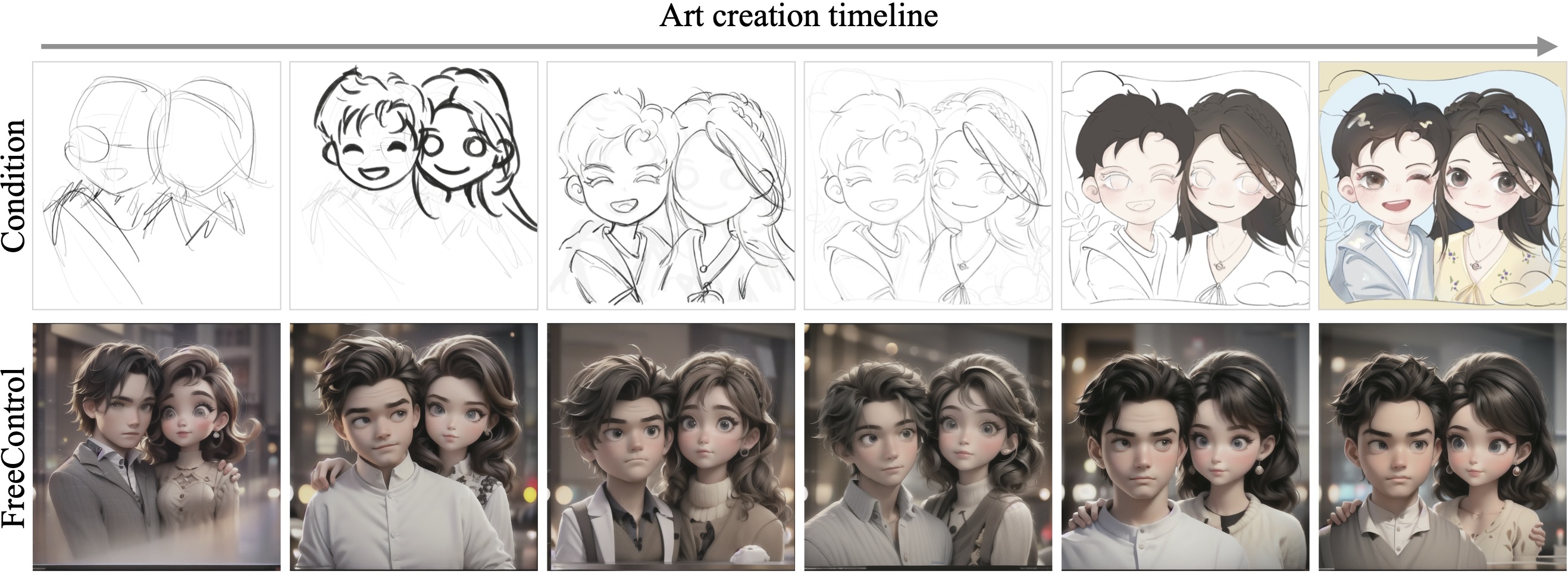}
\caption{\textbf{Controllable generation over the course of art creation.} Images are generated from the same seed with the prompt \textit{"a photo of a man and a woman, Pixar style"} with a customized model from \cite{civitai}. FreeControl yields accurate and consistent results despite evolving control conditions throughout the art creation timeline.}
\label{fig:continuous}
\end{figure*}

\smallskip
\noindent\textbf{Continuous control.}
Real-world content creation is a live experience, where an idea develops from a sketch into a more refined and finished piece of work. The intermediate states throughout this process may be interpreted as continuously evolving control signals. Figure~\ref{fig:continuous} illustrates how FreeControl may assist an artist in his or her content creation experience. It produces spatially accurate and smoothly varying outputs guided by constantly changing conditions, thus serving as a source of inspiration over the course of painting.

\begin{figure*}
\centering
\includegraphics[width=\linewidth]{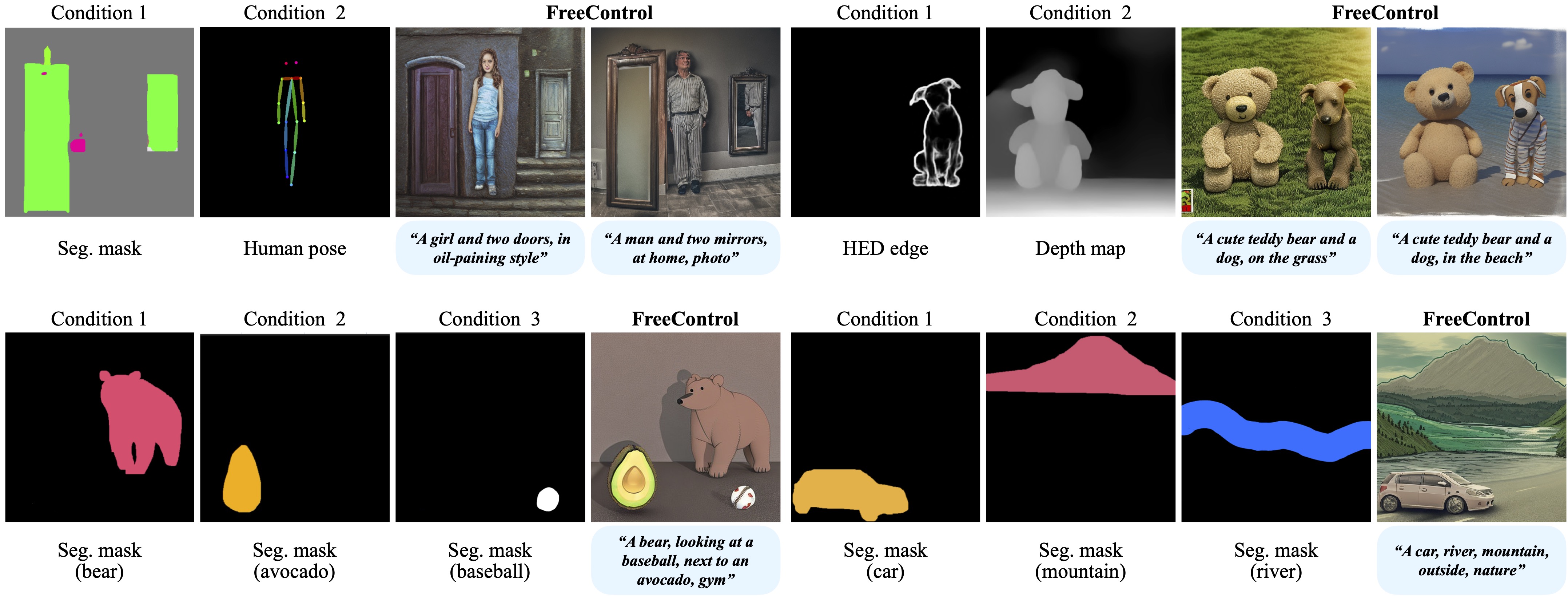}
\caption{\textbf{Qualitative results on compositional control}. FreeControl allows compositional control of image structure using multiple condition images of potentially different modalities.}
\label{fig:multi_gen}
\end{figure*}

\smallskip
\noindent\textbf{Compositional control.}
By combining structure guidance from multiple condition images, FreeControl readily supports compositional control without altering the synthesis pipeline. Figure~\ref{fig:multi_gen} presents our results using different combinations of condition types. The generated images are faithful to all input conditions while respect the text prompt.

\section{Additional Ablation Study}\label{supp:ablation}
We now present additional ablations of our model.

\begin{figure*}[htp]
\centering
\includegraphics[width=\linewidth]{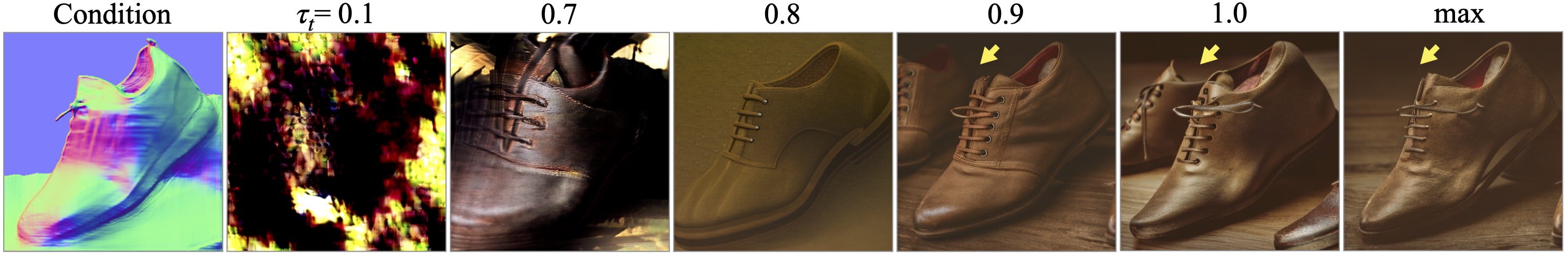}
\caption{\textbf{Ablation on threshold $\tau_t$.} Images are generated using the prompt \textit{"leather shoe on the table"}. Our dynamic threshold (max) encourages more faithful foreground structure and cleaner background in comparison to various hard thresholds (\eg, 0.1). 
}
\label{fig:ablation_trs}
\end{figure*}

\smallskip
\noindent\textbf{Choice of threshold $\tau_t$.}
Figure~\ref{fig:ablation_trs} demonstrates that no \emph{hard} threshold within the range of $[0, 1]$ can fully eliminate spurious background signal while ensure a foreground structure consistent with the condition image. By contrast, our \emph{dynamic} thresholding scheme, implemented as a per-channel \texttt{max} operation, allows FreeControl to accurately carve out the foreground without interference from the background.

\begin{figure*}[htp]
\centering
\includegraphics[width=\linewidth]{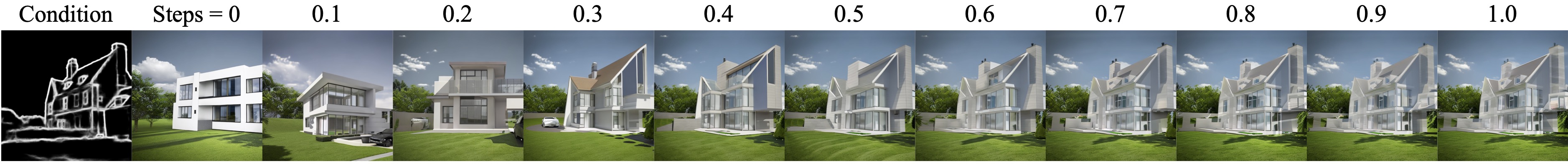}
\caption{\textbf{Ablation on number of guidance steps.} Images are generated using the prompt \textit{"a modern house, on the grass, side look"}. Applying guidance beyond the first $40\%$ diffusion steps (0.4) has little to no impact on the generation result.
}
\label{fig:ablation_steps}
\end{figure*}

\smallskip
\noindent\textbf{Number of guidance steps.}
Figure~\ref{fig:ablation_steps} reveals that the first $40\%$ sampling steps are key to structure and appearance formation. Applying guidance beyond that point has little to no impact on generation quality.

\begin{figure}[htp]
\centering
\includegraphics[width=\linewidth]{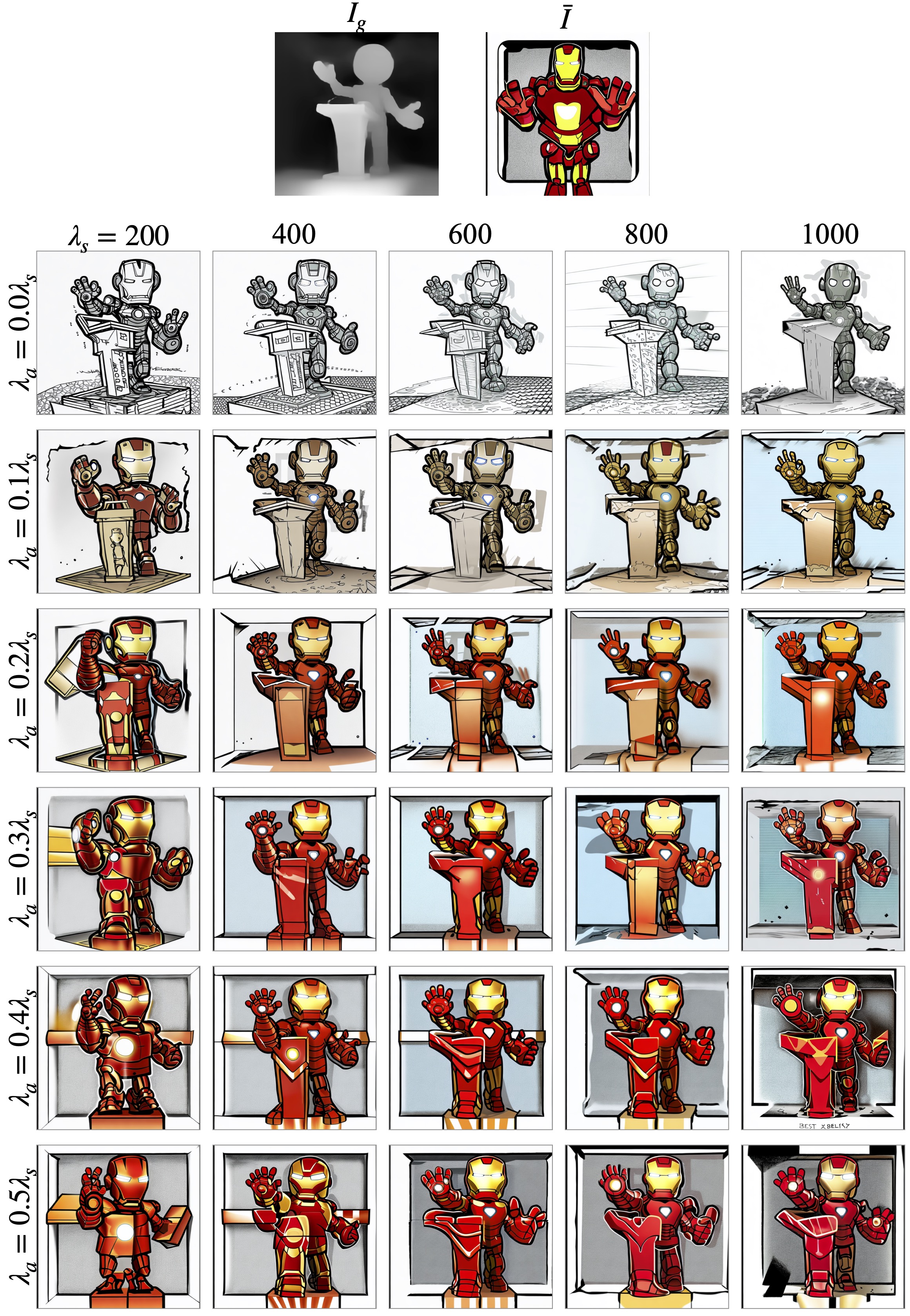}
\caption{\textbf{Ablation on guidance weights $\lambda_s$ and $\lambda_a$.} Images are generated with the prompt \textit{"an iron man is giving a lecture"}. FreeControl yields strong results across guidance weights.}
\label{fig:ablation_gs_ga}
\end{figure}

\smallskip
\noindent\textbf{Choice of guidance weights $\lambda_s$ and $\lambda_a$.}
Figure~\ref{fig:ablation_gs_ga} confirms that FreeControl produces strong results within a wide range of guidance strengths. In particular, the output images yield accurate spatial structure when $\lambda_s\ge400$ and rich appearance details when $\lambda_a\ge0.2\lambda_s$. We empirically found that these ranges work for all examples in our experiments.

\begin{figure}[htp]
\centering
\includegraphics[width=\linewidth]{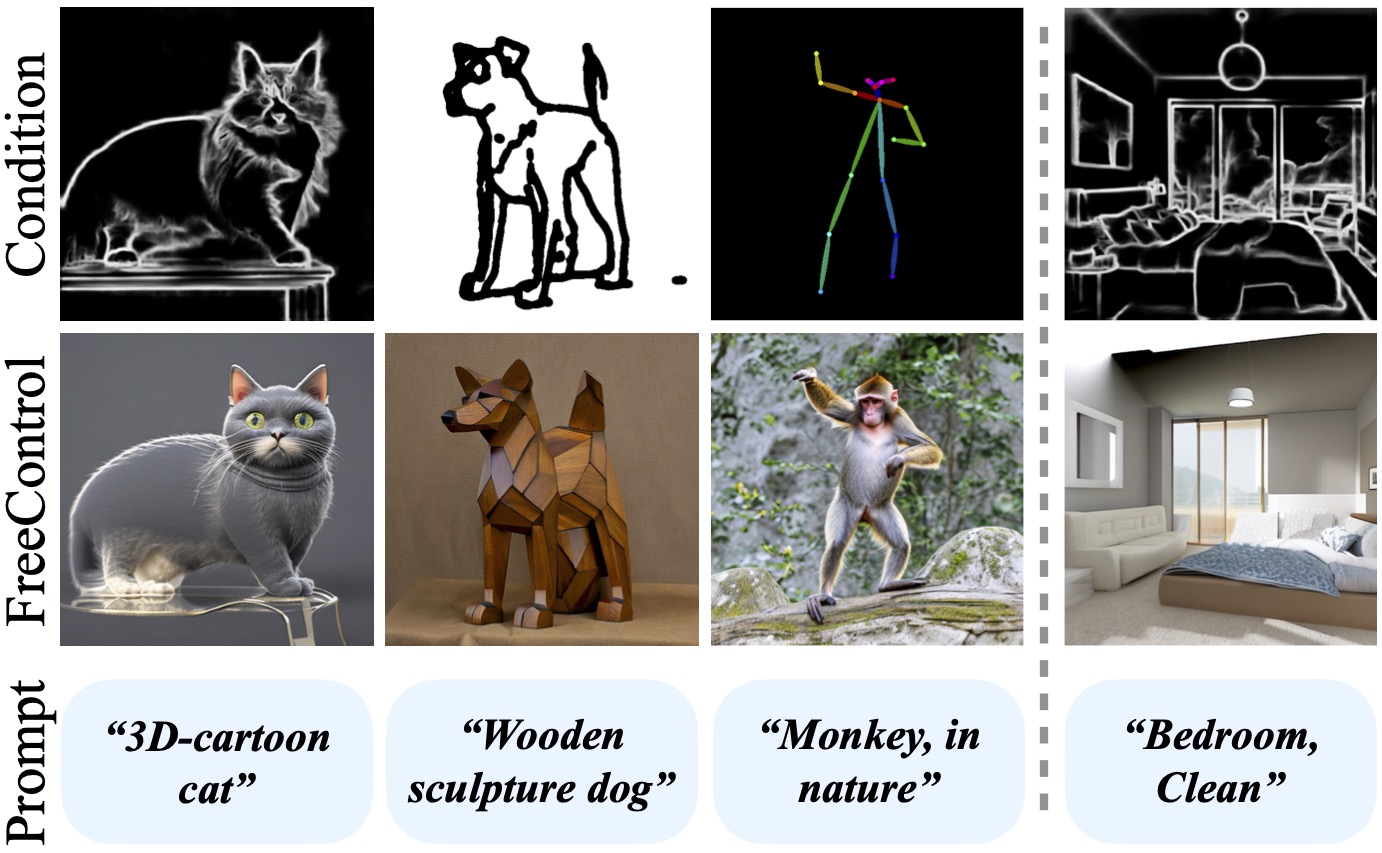}
\caption{\textbf{Ablation on basis reuse.} The semantic bases computed for \textit{"man"} enable the controllable generation of semantically related concepts (cat, dog and monkey) while fall short for unrelated concepts (bedroom). }
\label{fig:ablation_generalization}
\end{figure}

\smallskip
\noindent\textbf{Basis reuse across concepts.}
Once computed, the semantic bases $\mathbf{S}_t$ can be reused for the control of semantically related concepts. Figure~\ref{fig:ablation_generalization} provides one such example, where $\mathbf{S}_t$ derived from seed images of \texttt{man} generalize well on other mammals including \texttt{cat}, \texttt{dog} and \texttt{monkey}, yet fail for the semantically distant concept of \texttt{bedroom}.

\section{Limitations}\label{supp:limitation}

One limitation of FreeControl lies in its inference speed. Without careful code optimization, structure and appearance guidance result in $66\%$ longer inference time ($25$ seconds) on average compared to vanilla DDIM sampling~\cite{song2020ddim} ($15$ seconds) with the same number of sampling steps ($200$ in our experiments) on an NVIDIA A6000 GPU. This is on par with other training-free methods.

\begin{figure}[htp]
\centering
\includegraphics[width=\linewidth]{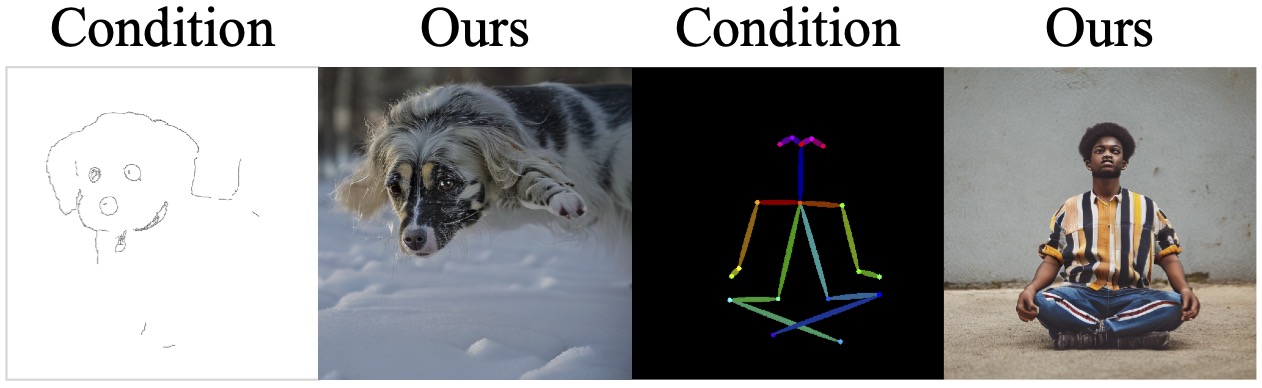}
\caption{\textbf{Failure cases.}
FreeControl does not anticipate missing structure in the condition image (\emph{left}) and may not accurately position fine structural details (limbs) in the output image (\emph{right}).
}
\label{fig:limitation}
\end{figure}

Another issue is that FreeControl relies on the pre-trained VAE and U-Net of a Stable Diffusion model to encode the semantic structure of a condition image at a low spatial resolution ($16\times16$). Therefore, it sometimes fails to recognize inputs with missing structure (\eg, incomplete sketch), and may not accurately locate fine structural details (\eg, limbs). Representative failure cases of FreeControl are illustrated in Figure~\ref{fig:limitation}.

\section{Societal Impact and Ethical Concerns}\label{supp:impact}
This paper presents a novel training-free method for spatially controlled text-to-image generation. Our method provides better control of the generation process with a broad spectrum of conditioning signals. We envision that our method provides a solid step towards enhancing AI-assisted visual content creation in creative industries and for media and communication. While we do not anticipate major ethical concerns about our work, our method shares common issues with other generative models in vision and graphics, including privacy and copyright concerns, misuse for creating misleading content, and potential bias in the generated content.

\clearpage

\clearpage

\clearpage

\newpage

\end{document}